# Fuzzy Constraints Linear Discriminant Analysis

Hamid Reza Hassanzadeh    Hadi Sadoghi Yazdi    Abedin Vahedian

Department of Computer Engineering, Ferdowsi University of Mashhad, Mashhad, Iran

Hamid_hassanzadeh2002@yahoo.com, sadoghi@sttu.ac.ir, vahedian@ferdowsi.um.ac.ir

***Abstract:*** *In this paper we introduce a fuzzy constraint linear discriminant analysis (FC-LDA). The FC-LDA tries to minimize misclassification error based on modified perceptron criterion that benefits handling the uncertainty near the decision boundary by means of a fuzzy linear programming approach with fuzzy resources. The method proposed has low computational complexity because of its linear characteristics and the ability to deal with noisy data with different degrees of tolerance. Obtained results verify the success of the algorithm when dealing with different problems. Comparing FC-LDA and LDA shows superiority in classification task.*

## 1.    INTRODUCTION

In recent years linear discriminant analysis has been applied to many practical problems dealing with classification of different data samples that are linearly separable. In [7] a cardiac arrhythmia diagnosis method is proposed using linear discriminant analysis on ECG signals. A new method for automatic hepatitis system based on linear discriminant analysis and adaptive network based on fuzzy inference system is proposed in [8]. In [9] a particle swarm optimization method is introduced for enhancing classification accuracy rate of linear discriminant analysis. In [10] an expert system on linear discriminant analysis and adaptive neuro-fuzzy inference system to diagnosis heart valve diseases is presented. Finally in [11] the authors have utilized a hybrid of fuzzy and LDA to recognize human continuous movements.

Since LDA was proposed for the first time, diverse variants has been proposed, which from our perspective, fall into two main categories; those that are directly reformed versions of LDA, [12,13] are of this category and those that are in fact the result of the fusion of LDA with other powerful tools such as fuzzy sets,[15] is of the like so is our paper.

In [15] different data samples are assigned different degree of importance according to how they are arranged in the feature space. The LDA in that paper tries to classify the data sets of different classes based on a relaxation in accomplishment of LDA objective function given the data samples with lower degree of importance the minor chance to be classified correctly.

The rest of this paper is organized as follows. In section 2 the LDA is explained. In section 3 our new fuzzy linear discriminator is introduced. In section 4 experimental results are presented. Finally in section 5 the discussions and conclusions are given.

## 2.    Linear Discriminant Analysis (LDA)

Linear Discriminant Analysis (LDA) is a method that seeks for weights of a linear combination of features that separate the training data samples which belong to different classes. The weights are then used in linear classifiers to make distinction between different samples of data pertaining to each of the classes. The binary linear classification problem as a special case of the more general form (the so-called multi-class classification) that is dealt with in this paper, tries to map each of the data samples to one of the two different classes $w_1$ and $w_2$.

A data set comprising of two classes is called linearly separable in the feature space X if there would be a function such as $g(x)$ in this space such that for every sample $x \in X$ we have

$$g(x) = w^T x + w_0 \begin{cases} > 0 \\ < 0 \end{cases} \Rightarrow x \in \begin{cases} w_1 \\ w_2 \end{cases} \quad (1)$$

where $w^T$ is the weight vector and $w_0$ is a threshold. In this case $g(x)$ is called the discriminator function which is the basis of linear classifiers. Equivalently $g(x)$ can be rewritten as

$$g(x) = v^T z \quad (2)$$

where $z = (1, x_1, x_2, ..., x_p)^T$ is the augmented pattern vector and $v$ is $(p+1)$-dimensional vector $(w_0, w_1, ..., w_p)^T$. Clearly and without loss of generality, one can redefine the samples in class $w_2$ by their negative values; according to this, (1) becomes

$$f(x) = v^T y > 0 \quad (3)$$

Where $y_i^T = (1, x_i^T), x_i \in w_1$ and $y_i^T = (-1, -x_i^T), x_i \in w_2$.

## 3. The Fuzzy Linear Discriminant Analysis (FC-LDA)

As opposed to the previous work on fuzzy LDA [15] where the data samples are assigned different degrees of importance, in this paper all the data samples are considered the same importance as it makes more sense; that is there shouldn't be any discrimination among different samples of a specific class since all of them are the members of the same group but with different values in different features. What is intended in this method is to find weights of linear discriminator which tries to maximize the objective function,$f$, subject to fuzzy constraints, in other words we are to

$$Maximize\ Obj \quad (4)$$
$$s.t.\ f(y_i) = v^T y_i \gtrsim 0$$

where $y_i$ are samples in $(p+1)$-dimensional space as stated in (3) for all $x_i \in w_1, w_2$.

The objective function in (4) might be any objective function depending on what criteria it is to be maximized. In our work we selected two different objective functions, the first is well known objective function called the perceptron criterion and the second is a linear objective function as stated in (5), (6) respectively.

$$Obj = \sum_{y_i \in Y} v^T y_i \quad (5)$$
$$Y = \{y_i | v^T y_i < 0\}$$

$$Obj = \sum_i v^T y_i \quad (6)$$

where $y_i$ are the data samples in $(p+1)$-dimensional space.

In other words, for (5) it is desired to minimize the number of misclassifications this yields a selection approach among the data samples which makes the objective function nonlinear. Contrary to (5) in (6) there will be no selection among data samples. Assuming the normal vector for the hyper plane discriminator,$v^T$, is unit vector, it is straightforward to see that the value of $Obj$ in (6) is the sum of signed distances from the reformed samples,$y_i^T$, to the discriminator and hence the maximum $Obj$ is regarded as the maximum overall distance from data samples to the discriminator, which means a larger noise margin we are subjecting to noisy data. This not only decreases the amount of computations but also makes the whole underlying problem a linear programming problem one as we will see shortly.

Putting all these together, we arrive at

$$Maximize\ c(v^T).v^T \quad (7)$$
$$Subject\ to\ (Av^T)_k \gtrsim b_k,\ k = 1,2,...,n$$
$$\|v^T\| = 1$$

where $v^T = (v_1, v_2, ..., v_{p+1})^T \in \mathbb{R}^n$ is the normal vector with unite magnitude for the hyper plane that we are seeking for, and

$c(v^T) = (c_1(v^T), c_2(v^T), \ldots, c_n(v^T))$ is the objective coefficient which may be either dependent or independent of variable $x$ depending on the choice we made for the objective functions stated in (5) and (6) respectively, and $A = [a_{ij}] \in \mathbb{R}^{n \times (p+1)}$ is the constraint matrix and finally the last constrain is a necessary non-linear condition so that the normal vector remains unit in magnitude when searching for an optimal solution. Disregarding the last constraint, we modeled the main problem as a fuzzy linear programming one for which c is regarded as $[(y^1_{s_1} + y^1_{s_2} + \cdots + y^1_{s_l}) \ldots (y^{p+1}_{s_1} + y^{p+1}_{s_2} + \cdots + y^{p+1}_{s_l})]$ where the trailing subscripts, $s_l$, are acquired according to what the choice of objective function will be, and finally A and $b_k$ as $\begin{bmatrix} y^1_1 & \cdots & y^{p+1}_1 \\ \vdots & \ddots & \vdots \\ y^1_n & \cdots & y^{p+1}_n \end{bmatrix}$ and $[0\ 0\ \ldots\ 0]^T$ respectively, where $y_i$ defined earlier. Note that the trailing superscripts and subscripts denote the dimensions and the sample numbers respectively. Note also that index k in (7) denotes the k'th constraint, $b_k$ is the resource pertaining to each constraint and n is the total number of constraints. Assuming $t_i$ be the tolerance of $i$'th resource, the fuzzy constraint related to the k'th resource is specified as

$$(Av^T)_k \le b_k + \theta t_k \quad (8)$$

where $\theta \in [0,1]$. Accordingly, the constraints in (7) could be restated as a fuzzy set with the following membership function,

$$\mu_k(v^T) = \begin{cases} 1 & (Av^T)_k < b_k \\ 1 - \frac{[(Av^T)_k - b_k]}{t_k} & b_k \le (Av^T)_k \le b_k + t_k \\ 0 & (Av^T)_k > b_k + t_k \end{cases} \quad (9)$$

$k = 1,2,\ldots,n$

Therefore the k'th constraint stated in (7) is now transformed to (9) which means for each of the them we are going to maximize the membership degree, $\mu_k(x)$, besides maximizing the objective criterion $cx$. In other words it is intended to find the solution to

$$\text{Maximize } cv^T, \mu_k(v^T)\ k = 1,2,\ldots,m \quad (10)$$

which is a multi-objective problem that searches for a solution for $m + 1$ objectives.

To solve the fuzzy linear programming problem stated in (7), we break it down into two crisp linear programming problems as follows

$$\text{Maximize } cv^T \quad (11)$$
$$\text{s.t.} \quad (Av^T)_k \le b_k, k = 1,2,\ldots,m$$
$$\|v^T\| = 1$$

and

$$\text{Maximize } v^T x \quad (12)$$
$$\text{s.t.} \quad (Av^T)_k \le b_k + t_k, k = 1,2,\ldots,m$$
$$\|v^T\| = 1$$

Solving the (11) and (12) gives the solutions $v^0$ and $v^1$ respectively. Assuming $z^0 = cv^0$ and $z^1 = cv^1$, we now define the following criterion for degree of optimality

$$\mu_0(v^T) = \begin{cases} 1 & v^T x > z^1 \\ 1 - \frac{z^1 - cv^T}{z^1 - z^0} & z^0 \le cv^T \le z^1 \\ 0 & v^T x > z^1 \end{cases} \quad (13)$$

That is the more $\mu_0(v^T)$ for the solution $v^T$, the better the degree of optimality will be disregarding how well the constraints are accomplished. As both the constraints and objective function are represented in the form of membership functions in (9) and (13), it makes sense to benefit max-min method to solving this multi-objective problem; specifically we have to find the solution to

$$\max_{\|v^T\|=1} \{\min[\mu_0(v^T), \mu_1(v^T), \ldots, \mu_m(v^T)]\} \quad (14)$$

This is equivalent to

$$\text{Maximize } \alpha \quad (15)$$
$$\text{s.t.} \quad \mu_0(v^T) \ge \alpha$$
$$\mu_k(v^T) \ge \alpha, i = 1,2,\ldots,m$$
$$\|v^T\| = 1, \alpha \in [0,1]$$

Inserting (13) and (9) into (15) we come to

$$\text{Maximize } \alpha \quad (16)$$
$$\text{s.t.} \quad cv^T \ge z^1 - (1-\alpha)(z^1 - z^0)$$

$$(Av^T)_k \le b_k + (1-\alpha)t_k , k = 1,2,\dots,m$$
$$\|v^T\| = 1 , \alpha \in [0,1]$$

Reforming (16) we arrive at

Maximize $c'v'^T$ (17)

s.t. $(A'v'^T)_k \le b'_k \quad k = 1,2,\dots,m$

$\|v^T\| = 1 , \alpha \in [0,1]$

where $v'^T = (\alpha, v_1, v_2, \dots, v_{p+1})^T$, $c' = (1,0,\dots,0)$, $A' = \begin{bmatrix} t_1 & y_1^1 & \cdots & y_1^{p+1} \\ \vdots & \vdots & \ddots & \vdots \\ t_n & y_n^1 & \cdots & y_n^{p+1} \\ z^1 - z^0 & c^1 & \cdots & c^{p+1} \end{bmatrix}$ and

$b'_k = (b_1 + t_1, \dots, b_n + t_n)^T$

which is absolutely a tractable linear programming problem. In other words having solved (16) the optimal solution subject to the fuzzy constraints is achieved.

## 4. Experiments and Results

We show the advantage of the proposed FC-LDA tool by comparing the noise margin and the $\alpha$ optimality criterion in two different data sets. For the sake of noise margin, we define the following quantities as noise margin criterion

$$NM_R = \frac{1}{\sum_{y_i \in Y_1} \frac{1}{v^T y_i}} \quad (18)$$

$Y_1 = \{(1, x_i^T) | x_i \in w_1 \}$

and

$$NM_R = \frac{1}{\sum_{y_i \in Y_1} \frac{1}{v^T y_i}} \quad (19)$$

$Y_2 = \{(1, x_i^T) | x_i \in w_2 \}$

Fig. 1 shows the first data set that is to be classified by means of our new method. Fig. 2 depicts the cross section of the linear separator in 3D that is achieved by ordinary linear discriminant analysis OLDA [16]. Figure 3,4 shows the result of the proposed algorithm based on the modified perceptron criterion and the perceptron criterion respectively, the degree of tolerance is set to 20% for both.

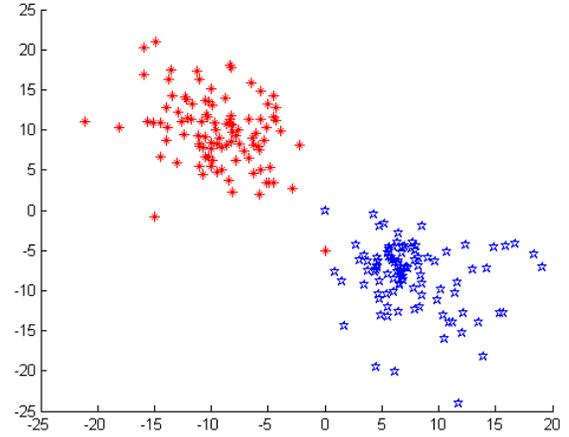

Fig 1. The first test data set

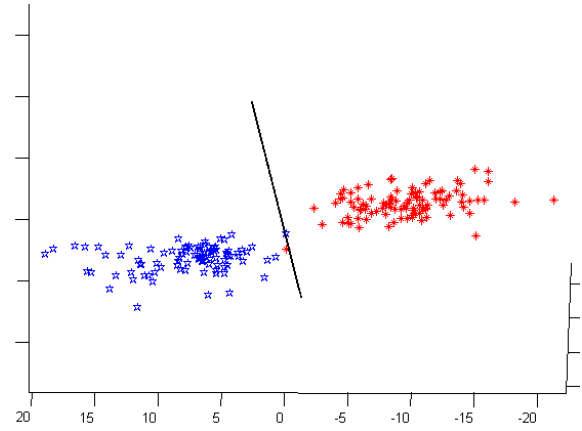

Fig 2. The cross section view of the OLDA hyper plane in 3D space

a)

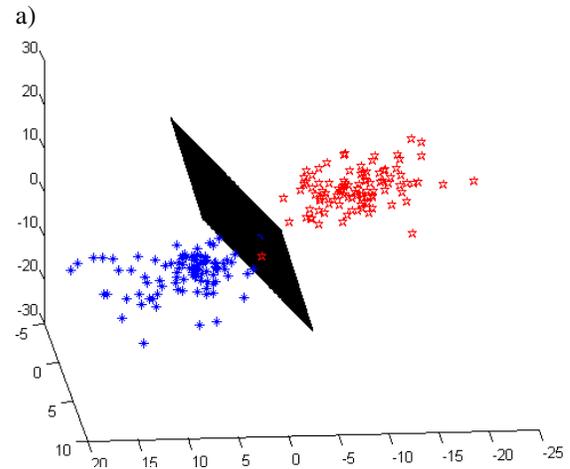

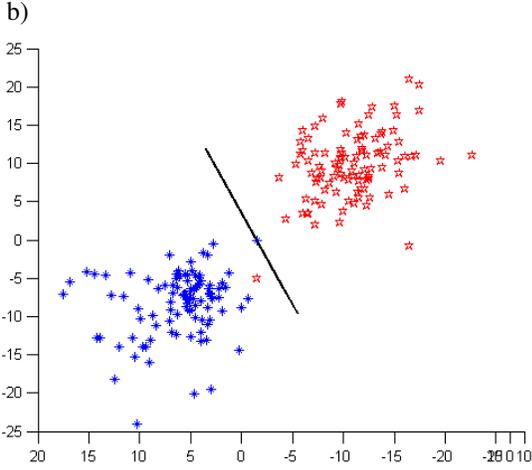

Fig 3. The results of the proposed based on the modified perceptron criterion in 3D space, $\alpha = 0.6691$ (a) the resulting discriminator hyper plane (b) the cross section view of the hyper plane

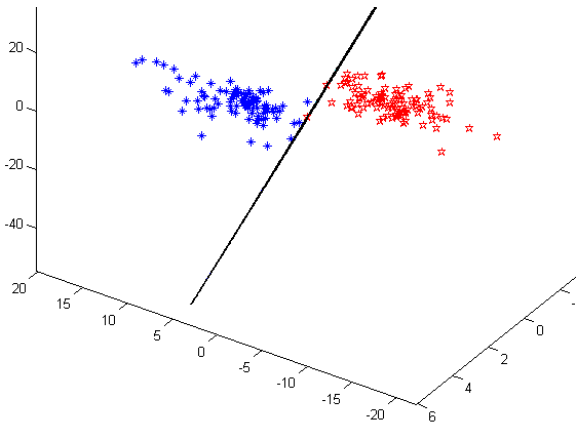

Fig 4. The cross section view of the resulting hyper plane based on perceptron criterion, $\alpha = 1$

The $\alpha$ corresponding to each of the figures are 0.6691 and 1.0 respectively. Notice how the perceptron criterion tries to minimize the misclassification error but at cost of decreasing the noise margin.

The second test is carried out on the famous iris data set [17] to classify the data pertaining to two different classes called Iris Versicolour and Iris Virginica. The selected features are sepal width and petal widths are both in cm. Fig, 5 shows the data set and Fig. 6 depict the cross section result of applying FC-LDA using perceptron criterion. Table 1, 2 also shows the calculated parameters for the tests already cited for two different tolerance degrees.

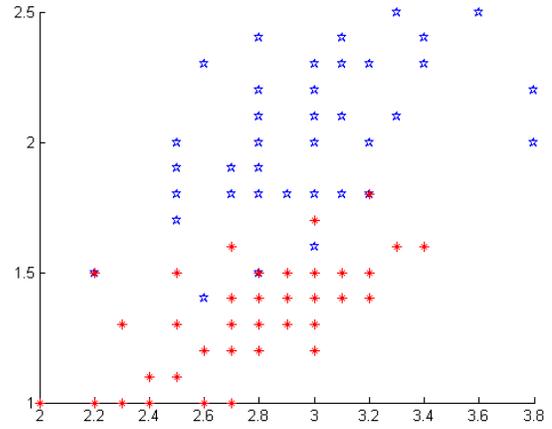

Fig 5. The second test data set

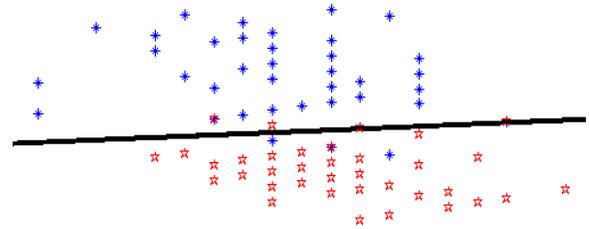

Fig 6. The cross section view of the resulting hyper plane based on perceptron criterion, $\alpha = 1$

Table 1. Parameters of the first data set

|  | Modified Perceptron Criterion | | Perceptron Criterion | |
|---|---|---|---|---|
| Tolerance | 10% | 20% | 10% | 20% |
| $\alpha$ | 0.5486 | 0.6691 | 1 | 1 |
| $NM_R$ | 0.1968 | 0.379 | 0.0505 | 0.0516 |
| $NM_L$ | 0.1170 | 0.1211 | 0.0642 | 0.0645 |

Table 2. Parameters of the second data set

|  | Modified Perceptron Criterion | | Perceptron Criterion | |
|---|---|---|---|---|
| Tolerance | 10% | 20% | 10% | 20% |
| $\alpha$ | 0.5018 | 0.5054 | 1 | 1 |
| $NM_R$ | 0.01070 | 0.0118 | $1.086 \times 10^{-7}$ | $7.51 \times 10^{-6}$ |
| $NM_L$ | 0.00048 | 0.00028 | $-1.86 \times 10^{-7}$ | $-7.43 \times 10^{-6}$ |

Table 1, 2 illustrates that the classifier based on the perceptron criterion has the best objective optimality but less noise margin which makes it undesirable in many cases, the modified perceptron criterion on the other hand falls short in objective optimality but has a better noise margin.

## 5. Conclusion

Preliminary applications prove that FC-LDA is a powerful and yet simple method for classifying diverse range of data sets. It is applicable in many real word problems that seems to be approximately linearly separable.

## 6-References

[1] Li-Xing Wang, A course in fuzzy systems and control, pp. 381-390, 1997

[2] F. Herrera , J.L. Verdegay, Three models of fuzzy integer linear programming, European Journal of Operational Research 83 (1995) 581-593

[3] Webb A. Statistical pattern recognition

[4] Luenberger D.G. Linear and nonlinear programming, pp.123-142, (2ed., AW, 1984)

[5] Werner, **B.** [1987], An interactive fuzzy programming system, Fuzzy Sets and Systems, 23, pp. 131-147.

[6] S.S. Wilks, Mathematical Statistics, Wiley, New York, 1962

[7] Yun-Chi Yeh , Wen-June Wang , Che Wun Chiou, Cardiac arrhythmia diagnosis method using linear discriminant analysis on ECG signals, Measurements 42 (2009) 778-789

[8] Esin Dogantekin, Akif Dogantekin, Derya Avci, Automatic hepatitis diagnosis system based on Linear Discriminant Analysis and Adaptive Network based on Fuzzy Inference System, Expert Systems with Applications, xxx not published yet (2009) xxxx

[9] Shih-Wei Lin , Shih-Chieh Chen, PSOLDA: A particle swarm optimization approach for enhancing classification accuracy rate of linear discriminant analysis, Applied Soft Computing, xxx not published yet (2009) xxx

[10] Abdulkadir Sengur, An expert system based on linear discriminant analysis and adaptive neuro-fuzzy inference system to diagnosis heart valve diseases, Expert Systems with Applications 35 (2008) 214–222

[11] Nikolaos Gkalelis, Anastasios Tefas, Ioannis Pitas, Combining Fuzzy Vector Quantization With Linear Discriminant Analysis for Continuous Human Movement Recognition, IEEE TRANSACTIONS ON CIRCUITS AND SYSTEMS FOR VIDEO TECHNOLOGY, VOL. 18, NO. 11, NOVEMBER 2008

[12] Shuiwang Ji, Jieping Ye, Generalized Linear Discriminant Analysis: A Unified Framework and Efficient Model Selection, IEEE TRANSACTIONS ON NEURAL NETWORKS, VOL. 19, NO. 10, OCTOBER 2008

[13] Yu-Wu Wang, An 0ptimal Linear Discriminant Analysis For Pattern Recognition, International Conference on Cyberworlds,2008

[14] Xiao-HongWu, Jian-Jiang Zhou, Fuzzy discriminant analysis with kernel methods, Pattern Recognition 39 (2006) 2236 – 2239

[15] Zeng-Ping Chen, Jian-Hui Jiang, Yang Li, Yi-Zeng Liang, Ru-Qin Yu, Fuzzy linear discriminant analysis for chemical data sets, Chemometrics and Intelligent Laboratory Systems 45_1999.295–302

[16] S.S. Wilks, Mathematical Statistics, Wiley, New York, 1962
[17] R.A. Fisher, Iris Plants Database, July 1988